\documentclass{article}

\usepackage{color}
\usepackage[nonatbib,final]{nips_2016}
\usepackage{setspace}
\usepackage{epsfig}
\usepackage[utf8]{inputenc} 
\usepackage[T1]{fontenc}    
\usepackage{hyperref}       
\usepackage{url}            
\usepackage{booktabs}       
\usepackage{amsfonts}       
\usepackage{nicefrac}       
\usepackage{microtype}      
\usepackage{amsmath,amsthm,amssymb}

\title{Learning Deep Neural Network Representations for Koopman Operators of Nonlinear Dynamical Systems}

\author{
    Enoch Yeung \\
  Data Science and Analytics Group\\
  Pacific Northwest National Laboratory\\
  Richland, WA 99354 \\
  \texttt{enoch.yeung@pnnl.gov} \\
  \And 
  Soumya Kundu \\ 
  Controls Group \\
  Pacific Northwest National Laboratory \\ 
  Richland, WA 99354 \\
  \texttt{soumya.kundu@pnnl.gov}\\
  \And
Nathan O. Hodas \\ 
  Data Science and Analytics Group \\
  Pacific Northwest National Laboratory \\
  Richland, WA 99354 \\
  \texttt{nathan.hodas@pnnl.gov} \\
 }

\begin{document}
\doublespacing
\maketitle

\begin{abstract}
 The Koopman operator has recently garnered much attention for its value in dynamical systems analysis and data-driven model discovery.   However, its application has been hindered by the computational complexity of extended dynamic mode decomposition; this requires a combinatorially large basis set to adequately describe many nonlinear systems of interest, e.g. cyber-physical infrastructure systems, biological networks, social systems, and fluid dynamics.  Often the dictionaries generated for these problems are manually curated, requiring domain-specific knowledge and painstaking tuning.  In this paper we introduce a deep learning framework for learning Koopman operators of nonlinear dynamical systems. We show that this novel method automatically selects efficient deep dictionaries, outperforming state-of-the-art methods.  We benchmark this method on partially observed nonlinear systems, including the glycolytic oscillator and show it is able to predict quantitatively 100 steps into the future, using only a single timepoint, and qualitative oscillatory behavior 400 steps into the future.     
\end{abstract}
\section{Introduction}
In 1931, B. O. Koopman published a paper showing that the evolution of any set of observables on a dynamical system can be expressed through the action of an infinite dimensional linear operator, the Koopman operator \cite{Koopman1931}.  Because the Koopman operator is a canonical representation of any autonomous dynamical system, in principle, its use can bring to bear linear analysis methods on nonlinear systems. 

The Koopman operator is especially powerful for inferring properties of dynamical systems that are either partially or completely unknown or that are too complex to express using standard methods in analysis.   Examples of such systems include biological networks, extremely large physical systems (which are intractable to analyze as white-box models), social networks, cyber-physical communication networks, and distributed computing systems that are subject to varying degrees of uncertainty.  For this reason, the Koopman operator has gained attention as an effective tool for data-driven model discovery.  The Koopman operator provides a data-driven model for comparing the asymptotic behavior of dynamical systems \cite{Mezic2005}, specifically as a function of its spectra \cite{Mezic2005, Rowley2009}.  Various algorithms have extended these methods using dynamic and extended dynamic mode decomposition, both for autonomous and controlled systems \cite{Williams2015, Brunton2016, Proctor2016}.   

The prevailing method for learning the Koopman operator from data is dynamic mode decomposition \cite{Rowley2009}.  Dynamic mode decomposition is the process of identifying a linear operator from temporally or spatially-linked data, ultimately with the objective of characterizing the spectrum of the operator.   There are many variants of dynamic mode decomposition, but the most recent advances in Koopman operator learning have emerged from {\it extended dynamic mode decomposition} \cite{Williams2015}.  

In extended dynamic mode decomposition, the idea is to lift the set of system observables from its native vector space into a higher dimensional space, usually through a nonlinear transformation \cite{Mezic2005, Rowley2009, Williams2015}.    When studying nonlinear systems, the hope is to lift the observables onto a nonlinear manifold, where there trajectories then evolve according to a linear law (the Koopman operator).   The set of nonlinear transformations is often referred to as a dictionary, while the choice of what dictionary functions to include is left to the discretion of the scientist, or based on {\it a priori} knowledge of the physical system of interest.     

\begin{figure}\label{fig:figure1}
\centering
\includegraphics[width=.8\columnwidth]{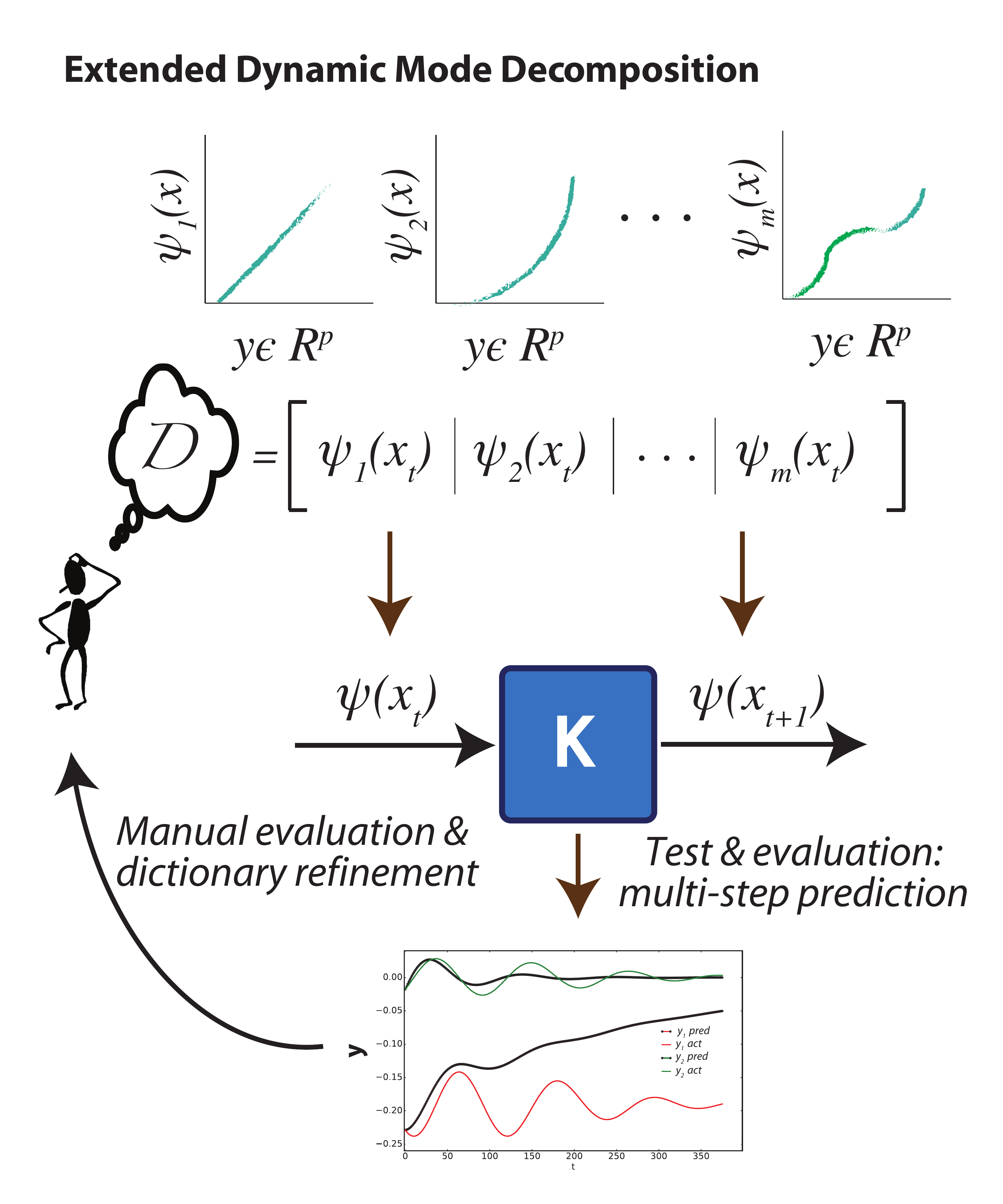}
\caption{A schematic illustrating the state-of-the-art approach for Koopman operator learning: extended dynamic mode decomposition.  The scientist postulates a variety of reasonable dictionary functions to include into the basis set, e.g. linear functions, orthonormal polynomial functions such as Hermite or Legendre polynomials, trigonometric functions, scalars, etc.   Once the dictionary is determined, the observables at each time point are mapped through the dictionary into a higher-dimensional space, where dynamic mode decomposition is performed to estimate a finite approximation of the Koopman operator.  If the model is not predictive, the user postulates new functions to include, based on intuition and domain specific knowledge of the problem.}
\end{figure}

Thus, extended dynamic mode decomposition allows for the study of complex nonlinear phenomena, e.g. the spectrum of nonlinear flows, but it is currently limited by a scientist's knowledge or intuition.   This inherently makes the learning process manual, since a dictionary must be intuited, implemented, and then evaluated, followed by careful review of the results by the scientist.  If the dictionary is not sufficiently complex, then the scientist must engineer prospective functions into the dictionary (see Figure \ref{fig:figure1}).  

In this paper we introduce a deep learning approach for training Koopman operators from data.   We show that deep neural networks can be used to both efficiently generate dictionaries and traverse function space during training, to obtain an accurate estimate of the Koopman operator.  We argue that deep neural networks lend themselves to larger problems than existing dynamic mode decomposition methods and highlight improved performance on a range of application problems.  Most importantly, we show how deep Koopman operators can be used to learn higher fidelity Koopman operator models, which improve on state-of-the-art multi-step prediction tasks (long-term forecasting).    We show how single-step prediction error can be misleading when evaluating dynamic mode decomposition (DMD) approaches and discuss the advantages of training dictionaries over one-shot computations (using DMD).   

The ideas in this paper are complementary to the recent work presented in \cite{Li2017}.   In \cite{Li2017}, the authors consider a 3-layer neural network with fixed depth and tanh activation functions to learn the dynamics of the Duffing oscillator and Kuramoto-Sivahinsky system.  Similarly we consider the use neural networks to dynamically update dictionaries for Koopman operator learning.  Additionally, our computational framework is implemented Tensorflow, allowing for direct manipulation of the objective functions, variations in the choice of activation functions such as RELUS, cRELUs, or ELUs over the entire network, variations in network depth, width, and composition, as well as direct access to deep training algorithms, e.g. AdaGrad \cite{Duchi2007}, ADAM \cite{Kingma2014} with hiearchical dropout \cite{Srivastava2014} for network regularization.  

\section{Deep Dynamic Mode Decomposition for Koopman Operator Learning}
\subsection{Extended Dynamic Mode Decomposition} 
We consider a discrete-time nonlinear dynamical system of the form 
\begin{equation}
\begin{aligned}
x_{n+1} = f(x_n) \\
y_n = h(x_n)
\end{aligned}
\end{equation}
where $f \in \mathbb{R}^n$ is continuously differentiable and $h \in \mathbb{R}^p$ is continuously differentiable.   The function $f$ is the state-space model and the function $h$ maps current state values $x_n\in\mathbb{R}^n$ to a vector of  {\it observables} or outputs $y_n \in \mathbb{R}^p.$    The Koopman operator of this system must satisfy the equation
\begin{equation}\label{eq:singleupdate}
\begin{aligned}
\psi(x_{n+1})&= {\cal K}(\psi(x_n))\\
y_n &= H(\psi(x_n))
\end{aligned}
\end{equation}
where $\psi(x_n) \in \mathbb{R}^m$ is an observable function, where $m \leq \infty$, that defines the lifted space of observables.  In theory, $\psi$ must belong to a space of observables under which $\psi(x_n)$ is ${\cal K}$-invariant for all $n$.  This ensures that the Koopman operator comprehensively describes the flow of the observable trajectory $(x_1, x_2, ... ).$     More precisely, we suppose that the span of the dictionary matrix 
\begin{equation}{\cal D} \equiv \Psi(y) = \begin{bmatrix} \psi_1(x) & \psi_2(x) & \hdots  \end{bmatrix}
\end{equation}   
contains all observables $\psi(x_n)$ for all $n$.  

The challenge, in practice, is that the dictionary matrix ${\cal D}$ is unknown.  This makes it difficult to ascertain what and how many functions to include, let alone the minimal number of functions, to ensure ${\cal K}$-invariance.   In reality, the data is provided and set as a series of points $(x_1,...,x_N)$, so whatever functions picked must satisfy the invariance property with respect to the set 
\[ \{\psi(x_1), ...., \psi(x_N)\}.\]   
In extended dynamic mode decomposition, we use an expansive set of orthonomal polynomial basis functions.  However, this approach does not scale well and suffers from overfitting with an increasing number of dictionary functions.  Each element in $x_k, k = 1,...,n$ is lifted to $m$ polynomials, making for $mn$ distinct dictionary functions.  These dictionary functions are often monomial, so quadratic or cubic cross terms are computed to model nonlinear interaction between states, which further exponentiates the size of the dictionary matrix.  Finally, since the Koopman operator is unknown, it is generally presumed to be fully parameterized and gradually regularized during optimization.  This makes it difficult to solve problems with even only 10-20 outputs, since the Koopman operator will quickly grow to contain thousands of potential entries. Thus, it is critical to regularize the learning problem, to minimize overfit.  For dynamic mode decomposition of a Koopman operator, the learning problem is typically cast as follows.  

Let $\psi$ be an observables vector defined on $y_n$, for $n = 1,..., N$.  We stack the observables in two matrices $Y_f$ and $Y_p$ 
where \begin{equation} \begin{aligned}
\begin{array}{ccc}
Y_f  =  \left[\begin{array}{c|c|c}\psi(x^{(0)}_{n+1})  &  \hdots  &  \psi(x^{(0)}_1) \\ \vdots  &  \ddots  &  \vdots \\ \psi(x^{(p)}_{n+1}) & \hdots & \psi(x^{(p)}_1 )  \end{array} \right]  &, \mbox{\hspace{3mm}} 
Y_p   =  \left[\begin{array}{c|c|c}\psi(x^{(0)}_{n})  &  \hdots  &  \psi(x^{(0)}_0) \\ \vdots  &  \ddots  &  \vdots \\ \psi(x^{(p)}_{n}) & \hdots & \psi(x^{(p)}_0 )  \end{array} \right]. \end{array}
\end{aligned} \end{equation}
From these definitions, it is straightforward to horizontally stack the Koopman equation for a {\it single observable update} (\ref{eq:singleupdate}) to obtain
\begin{equation}
Y_f = K Y_p. 
\end{equation}
The $L_1$-regularized version of this problem is 
\begin{equation}
|| Y_f - K Y_p ||_2  + \lambda ||K||_{2,1} 
\end{equation}
where the second term is computed first as the 2-norm of each column vector, followed by a 1-norm on all of the 2-norms, to encourage sparsity.   The parameter $\lambda$ is a hyper-parameter for tuning the sparsity constraint.   We remark that the problem formulated here is the transpose of the problem formulated in \cite{Williams2016} and is convex.  This formulation thus is guaranteed to converge to a unique global estimate of the Koopman operator, so long as it is computationally tractable.  Our subsequent experimental results will draw comparisons against this state-of-the-art approach for estimating Koopman operators. 

\begin{figure}[]
\centering
\includegraphics[width=1.0\columnwidth]{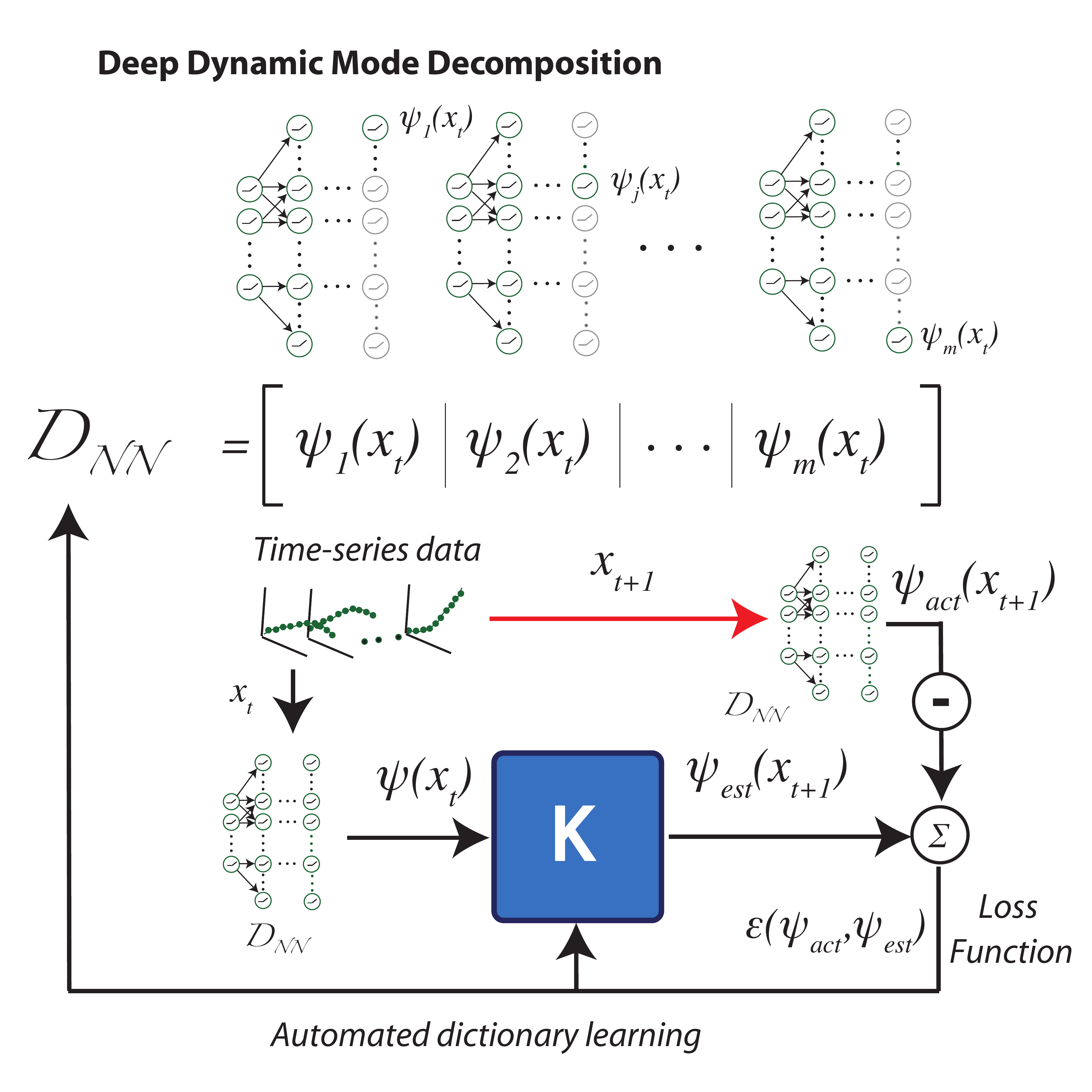}
\caption{A schematic illustrating our deep Koopman operator learning approach.  The entire dictionary is specified as the output or outputs of a deep artificial neural network.  The deep neural network can be a multi-layer feedforward network, a convolutional network, a recurrent network etc. The observables are forward propagated through the neural network to achieve lifting.  The prediction error is then used to compute the loss function and refine the dictionary automatically.     }
\end{figure}
 
\subsection{Deep Dynamic Mode Decomposition}
In deep dynamic mode decomposition, we define the dictionary as the output of a deep neural network.  The deep neural network can be specified to have any architecture upstream, so long as it takes $y_k$ as an input.      The advantage of employing deep neural networks is that they have rich expressivity \cite{Raghu2016, Bengio2011}, can be trained automatically,  and scale to extremely large networks \cite{Bengio2013}.    With each additional layer, the number of new nonlinear functions grows combinatorially, but only with a linear increase in the model parameters.    We setup the deep Koopman operator learning problem as follows.   Define the deep dictionary as  
\begin{equation}
{\cal D}_{NN} = N_\Psi(x_n)
\end{equation}
where $N_\Psi(x_n)$ is a neural network of choice.  The Koopman operator problem is cast similarly as with extended dynamic mode decomposition:
\begin{equation}
\begin{aligned}
\min_{K, \theta} ||N_\Psi(x_{n+1},\theta) & - K  N_{\Psi} (x_{n},\theta)||_2 + \lambda_1 ||K||_2  + \lambda_2 ||(\theta)||_1 
\end{aligned}
\end{equation}
 where 
 \[ 
 \theta = (W_1,...,W_{d}, b_1, ..., b_d ,...)
 \]
are the numerical parameters of the neural network.  For example, in the case of a multi-layer deep feedforward residual network, 
 \begin{equation}N_\Psi(x_{n}) = h_d\circ h_{d-1} \circ \hdots \circ h_{1}(x_{n})\end{equation} with $h_j(h_{j-1}) = \sigma_j(W_{j-1} h_{j-1} + b_{j-1}),$ and $W_{j} \in \mathbb{R}^{r_j \times r_{j-1}}$.   The parameters $d, r_j$ are hyperparameters that are optimized using a combination of dropout and regularization, while the the activation function $\sigma_j$ can be chosen to be one of the many activation functions that result in fast convergence, e.g. ELU, cRELU, or the RELU.

Given this formulation, we remark it is also possible to initialize a neural network, akin to selecting a randomly parameterized dictionary, and directly solve the dynamic mode decomposition problem.  Computing $Y_f$ and $Y_p$ as before, yields 
\begin{equation}
\begin{aligned}
\begin{array}{ccc}
Y_f  =  \left[\begin{array}{c|c|c}N_\Psi(x^{(0)}_{n+1})  &  \hdots  &  N_\Psi(x^{(0)}_1) \\ \vdots  &  \ddots  &  \vdots \\ N_\Psi(x^{(p)}_{n+1}) & \hdots & N_\Psi(x^{(p)}_1 )  \end{array} \right]  &, \mbox{\hspace{3mm}} 
Y_p   =  \left[\begin{array}{c|c|c} N_\Psi(x^{(0)}_{n})  &  \hdots  &  N_\Psi(x^{(0)}_0) \\ \vdots  &  \ddots  &  \vdots \\ N_\Psi(x^{(p)}_{n}) & \hdots & N_\Psi(x^{(p)}_0 )  \end{array} \right]. \end{array}
\end{aligned} \end{equation}
However, we observed in practice that this approach did not produce accurate estimates of the Koopman operator.   The key difference between deep Koopman operator learning  and dynamic mode decomposition is that in deep Koopman operator learning, the dictionary and the Koopman operator are learned simultaneously.  This increases the number of degrees of freedom in the problem, but no more than was initially present.  In reality, the dictionary functions {\it and} the Koopman operator are unknown. Dynamic mode decomposition simplifies the problem  by having the user impose a particular choice of dictionary, followed by direct optimization of the Koopman operator assuming a fixed dictionary.   However, both (sets of) variables are usually unknown in data-driven applications. 

Finally, we remark that once the Koopman operator is obtained, as with dynamic mode decomposition, an eigendecomposition of $K$ will yield the eigenmodes for the neural network basis functions.  The weighted combination of those basis functions will define the Koopman eigenfunctions, with corresponding Koopman eigenvalues in the diagonalization or Jordan form of $K$. 

Which neural network is used ultimately impacts the nature of the dictionary embedding the observables. If the neural network has memory, e.g. an LSTM network or a recurrent neural network, then the dictionary functions can be viewed as time or space varying.   Their properties change according to some memory state, which is dynamic from one time point $x_n$ to the next $x_{n+1}$.    If the network has hybrid or bifurcative behavior, e.g. a switching network, then the dictionary functions may have the capacity to model multiple invariant subspaces of the Koopman operator simultaneously.  The challenge is knowing when the dictionary functions are trained accurately to recover the true Koopman operator.  In reality, unless the system can be Carleman linearized \cite{Kowalski1991}, it is difficult to ascertain the complete dictionary.   An empirical measure for the fidelity of the Koopman operator is its forecasting or prediction accuracy.  In the next section, we review a collection of several experiments performed to 1) gain insight into how deep Koopman operators learn basis functions, 2) show the ability of deep Koopman operators to learn completely unknown governing laws for a biological network and forecast multiple steps into the future, and 3) demonstrate the ability of deep Koopman operators to learn an approximate coarse-grained model for a power transmission system.


\section{Experimental Results}
\subsection{ The Emergence of Koopman Basis Functions During Training}
As a first experiment, we trained deep and classical Koopman operators on randomly generated linear systems with partially observed dynamics. When linear systems have partial state output, their Koopman operator is a coarse grained model of the underlying latent dynamics.   For discrete time linear systems, a closed form expression for the Koopman operator exists, called the dynamical structure function \cite{Goncalves2008}.   The transfer function entries within a dynamical structure function act as linear fractional transformations on the input signals.   Estimating the Koopman operator in this scenario requires discovering the basis functions (in the time-domain) for those transfer function entries. 

We first performed experiments on a range of small networks, ranging from 4-128 nodes.  A subset of the nodes were allocated as ``latent" during the experiment.   The systems were simulated in Python and time-series data was collected and divided into training and test data.  For optimization, we used the AdaGrad algorithm, as implemented in Tensorflow.    We tried randomly shuffling and varying batch size; we found that shuffling did not have a significant effect on the performance of the deep Koopman operator.  As expected, if batch size became too small, e.g. less than 10 samples per batch, the algorithm would not converge. All training and validation was done using Python Tensorflow 1.0, on an NVIDIA TitanX Pascal or P40 system.  

The results of the study are presented in Table \ref{table:rand-osc-table}.  The deep Koopman operator had on average  $1\%$ one-step training error and one-step test error after 10,000 iterations (see Table 1). Further iterations reduced this error down to $0.1\%.$  In contrast, the E-DMD Koopman operator averaged over $10\%$ on all network sizes.        It was also difficult to execute regularized extended dynamic mode decomposition on larger dynamical systems.  The primary challenge is that the convex optimization routines run out of memory.  We tested the algorithm both on a NVIDIA GPU TitanX Pascal system, as well as a NVIDIA P40 system.  For moderate sized networks with more than 10 variables, the quadratic and cubic terms induces rapid expansion in the size of the Koopman operator, which ultimately slowed the CVXPY solvers.    
\begin{table}[t]
  \caption{Prediction Accuracy of E-DMD vs Deep-DMD Koopman Operators on Partially Observed Linear Systems (POLS) and the Glycolytic Pathway }
  \label{table:rand-osc-table}
  \centering
  \begin{tabular}{lllll}
    \toprule
    \cmidrule{1-2}
    Network Size \\(POLS)    & E-DMD  Train. Err. & E-DMD Test Err.    &  D-DMD Train Err. & D-DMD Test Err. \\
    \midrule
    4 & 14.5\% &  14.5\%  &   <1\% & <1\%\\
    8  & 15.10\% &   15.10\%&   <1\%& <1\%  \\
    16 & 14.3\% & 15.2\% &<1\%&<1\% \\ 
    32 & 11.0\% & 11.8\%& <1\%& <1\%\\ 
    64 & 14.5\% & 16.4\%& <1\%& <1\%\\ 
    128 & 12.5\% & 12.5 \%& <1\% &<1\%\\ 
        \midrule
    Species Measured & E-DMD Train. Err & E-DMD Test Err. & D-DMD Train Err. & D-DMD Test Err. \\
      7 &   2.67\%  & 2.5\%    & 0.5\%    & 0.5\% \\
      5 &   4.95& 5.2\%  \%    & .47\% & .66\% \\
      3 &       4.67\% & 5.05\%& .76\% & .84\% \\
    \bottomrule
  \end{tabular}
\end{table}
On the other hand, we were able to compute the deep Koopman operator for increasingly higher dimensional systems.  For small systems, a network width of 20 nodes and a depth of 3-5 layers was sufficient.  With networks ranging from 10-100 states, we found the deep Koopman operator could be trained if the neural network was sufficiently deep.  This may be because the deeper the neural network, the more efficiently it encodes complex nonlinear functions.  Once again, the one-step training error and test error were approximately $1\%$.    
\begin{figure}\label{fig:figure3}
\centering
\includegraphics[width=1.0\columnwidth]{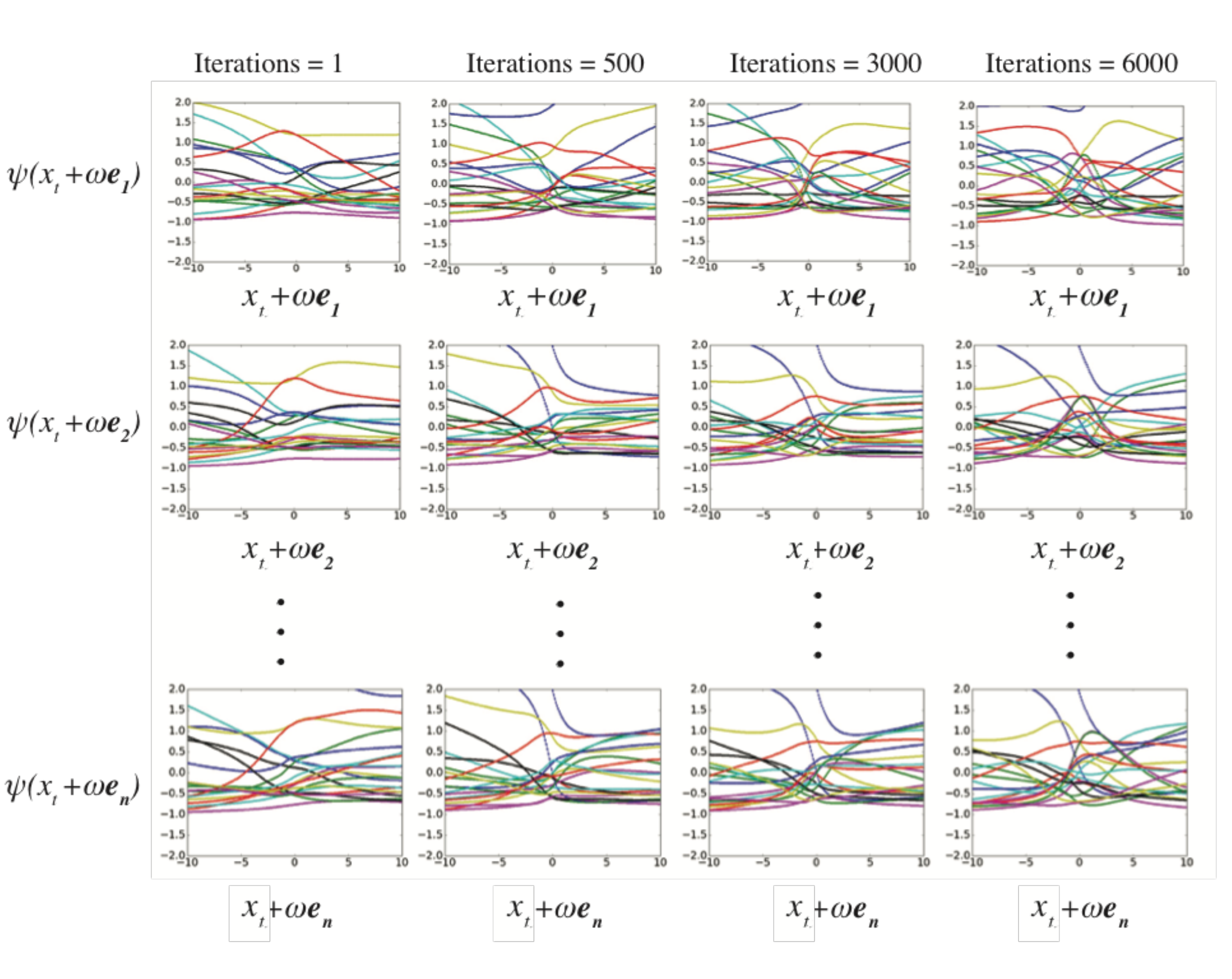}
\caption{The Koopman basis functions generated by a 20 layer drop-out ELU neural network. The x axis shows the size of the perturbation to a particular dimension of the observable vector $y_n$ and the y axis shows the response of the basis functions.}
\end{figure}
To better understand what the deep Koopman operator was learning, we visualized the response of the neural network dictionary functions to basis perturbations along each observable axis.   We found that much of the drastic changes in the basis function profile occurs in the first 10,000 iterations of training.  During this process, the basis functions initialize as flat constant functions across all points and gradually train to spike, dip, or ramp in response to input signals (see Figure 3).   This allows our Koopman dictionary to be updated during training, while simultaneously optimizing the Koopman operator.   This may explain why the deep Koopman operator is able to achieve a higher fidelity than the extended DMD Koopman operator.


\subsection{Learning A Koopman Operator for the Glycolysis Pathway}
The ultimate benchmark of whether or not a Koopman operator has been successfully learned is if the operator is  able to predict multiple points into the future, using only a single root time-point.   Any small amount of error in the Koopman operator estimate will propagate and compound on itself.    Moreover, the ultimate challenge is to learn the Koopman operator on a nonlinear system for which there is no known Carleman linearization or Koopman invariant subspace.  

The glycolysis pathway is a fundamental pathway to metabolism in biology.  We use the standard model introduced in \cite{Daniels2015}, a seven state nonlinear model with Michaelis-Menten dynamics.  It is given as follows: 
\begin{equation}
\begin{aligned}
\frac{d S_1}{dt} &= J_0 - \frac{k_1 S_1 S_6} {1+ \left( S_6/K_1 \right)^q}, \\ 
\frac{d S_2}{dt} &= 2 \frac{ k_1 S_1 S_6} { 1+ \left(S_6 / K_1\right)^q} - k_2 S_2 (N-S_5) - k_6 S_2 S_5, \\ 
\frac{ d S_3}{dt} & = k_2 S_2 (N-S_5) - k_3 S_3 (A-S_6) , \\
\frac{ d S_4}{dt} & = k_3 S_3 (A- S_6) - k_4 S_4 S_5 - \kappa ( S_4 - S_7) ,\\ 
\frac{d S_5}{dt} & = k_2 S_2 (N-S_5) - k_4 S_4 S_5 - k_6 S_2 S_5 , \\ 
\frac{d S_6}{dt} &= -2 \frac{k_1 S_1 S_6}{ 1 + (S_6/K_1)^q} + 2 k_3 S_3 ( A- S_6) - k_5 S_6 ,\\ 
\frac{d S_7}{dt} &= \mu \kappa (S_4 -S_7) - k S_7,
\end{aligned}
\end{equation}
\begin{figure}
\includegraphics[width=\columnwidth]{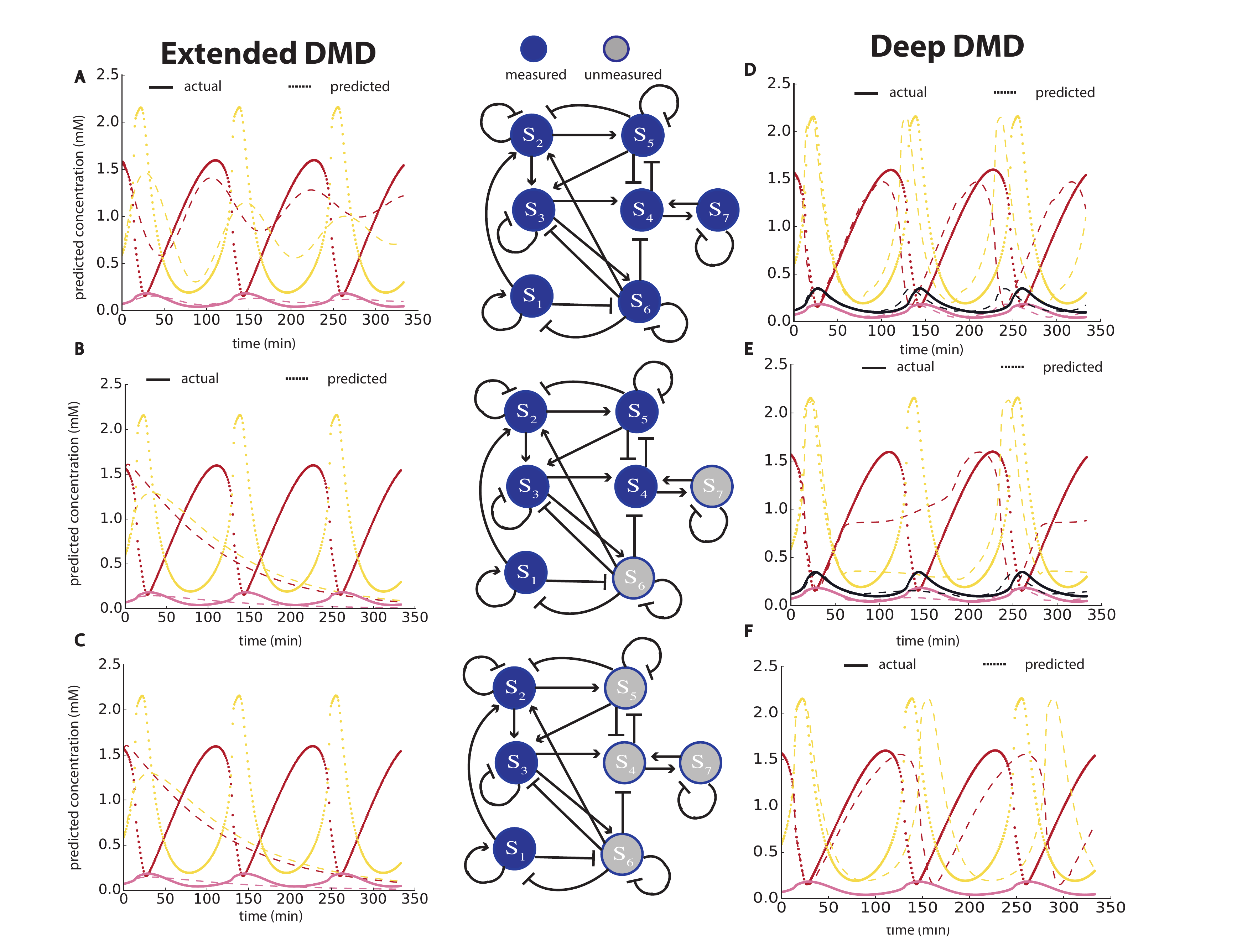}
\caption{ Multi-step prediction using a single time-point  on the glycolytic oscillator, trained using measurements from {\bf A}) 7 states with extended dynamic mode decomposition (E-DMD),   {\bf B}) 5 states with E-DMD,  {\bf C}) 3 states with E-DMD, {\bf D}) 7 states with our deep dynamic mode decomposition algorithm (D-DMD), {\bf E}) 5 states with D-DMD, {\bf F}) 3 states with D-DMD.}
\end{figure}
The network has 19 edges and 7 nodes, autoregulatory feedback, both positive and negative feedback.  We consider a challenging scenario, where the network parameters are tuned to oscillate.  As seen in Table 1, the deep Koopman operator achieved less than $.1\%$ error on all test cases, whether $n=7,5,$ or $3$ nodes in the glycolysis network were measured.  Interestingly, the deep Koopman operator is able to predict multiple time-steps into the future ($\sim$ 100 timepoints) with quantitative accuracy and qualitative accuracy for the full duration of the simulation (500 timepoints).   

\subsection{Learning Koopman Operators on Power Systems}
Our next example is a multi-machine power systems network \cite{Kundur1994}. There are multiple time-scales of dynamics involved in power systems; {\it transient dynamics} represent the fastest of all time-scales.  Transient instability often leads to instability in frequency and voltage levels, which can result in cascading blackouts.  Transient dynamic models are often calibrated, but small fluctuations in parameters, as well real-time state uncertainty is often unknown.  However, measurement data for rotor speed and angle is often available, motivating the use of data-driven methods for discovering real-time relationships between power flow variables and swing dynamics.  In addition, discovering a Koopman operator from data enables direct application of linear analysis methods for contingency planning \cite{Fouad1991}, which has traditionally been addressed using brute-force simulation studies \cite{Huang2009}.


   More specifically, we are interested in power systems transient dynamics, which models the evolution of rotor angles of the synchronous machines at sub-second to a couple of seconds timescales. At this timescale, the dynamics of interest can be modeled as the classical swing dynamics:
\begin{subequations}
\begin{align}
\forall i\in\lbrace 1,2,\dots,n\rbrace:\quad\dot{\delta}_i&=\omega_i\,,\\
\dot{\omega}_i&=\frac{1}{M_i}\left(-D_i\,\omega_i+P_{m,i}-P_{e,i}\right)\,,\\
\text{where }\,P_{e,i}&=V_i\sum_{j=1}^nV_j\left(G_{ij}\cos(\delta_i-\delta_j)+B_{ij}\sin(\delta_i-\delta_j)\right)\,.
\end{align}
\end{subequations}
$\delta_i$ and $\omega_i$ represent the generator rotor angle and speed, respectively. Rotational inertia ($M_i$) and damping ($D_i$) are the parameters associated with each synchronous machine, while $P_{m,i}$ is its mechanical power input. $P_{e,i}$ is the network term that represents the total electrical power output from the generator terminal into the power grid. In the Kron-reduced network representation (each node in the network is a machine), the electrical output is a sum of the total power flowing from the generator to all other generators connected to it. Power flowing between two generators is a function of their voltages ($V_i,\,V_j$), rotor angles ($\delta_i,\,\delta_j$) and the transfer conductance ($G_{ij}$) and susceptance ($B_{ij}$). The values of $G_{ij}$ and $B_{ij}$ are zero if there is no power line between $i$ and $j$. Note that since only the rotor angle difference (and not their absolute values) matter, It is generally customary to use a generator angle as the reference angle, and express all the rotor angles relative to the reference angle.

The perturbed system dynamics, assuming a disturbance $\Delta_i$ to the initial condition  of the relative angle $\delta_i(0)$ is modeled as 
\begin{subequations}
\begin{align}
\forall i\in\lbrace 1,2,\dots,n\rbrace:\quad\dot{\delta}_i&=\omega_i\ + (1) \Delta_i \delta(t) = \omega_i \ + 1 u_i(t),\\
\dot{\omega}_i&=\frac{1}{M_i}\left(-D_i\,\omega_i+P_{m,i}-P_{e,i}\right)\,,\\
\text{where }\,P_{e,i}&=V_i\sum_{j=1}^nV_j\left(G_{ij}\cos(\delta_i-\delta_j)+B_{ij}\sin(\delta_i-\delta_j)\right)\,.
\end{align}
\end{subequations}

We next explored the use of deep Koopman operators to learn the dynamics of a IEEE 39 bus benchmark system.  The system has 39 buses and 10 generators; it is a classical model system that has been well studied in the context of coherency and spectral modes, especially using the Koopman operator \cite{MezicPower}.    For a more detailed exposition on the relationships between spectral stability of a data-driven  Koopman operator and that of the underlying system, we refer the reader to \cite{MezicSpectral,MezicSpectralStability}. 

In our numerical experiments, we sought to characterize the ability of deep dynamic mode decomposition to learn the transient dynamics of a power system over a wide range of initial conditions.  For training, we generated 1000 randomly generated trajectories of the swing dynamics.   The same training data was provided to both deep and extended dynamic mode decomposition algorithms.  For extended dynamic mode decomposition, we used Legendre basis polynomials \cite{Williams2015}, using L1 regularization to find the best scoring Koopman operator (in terms of 1-step prediction training error).  We iterated over a range of maximum polynomial degrees, selecting the best scoring Koopman operator.  Our choice of Legendre polynomials was based on the recommendations of \cite{Williams2015}.  We remark that for the IEEE 39 bus benchmark system, it may be possible that a better choice of bases could be thin-plate radial basis functions (RBFs) or Hermite polynomials. 

\begin{figure}
\centering
\includegraphics[width=\columnwidth]{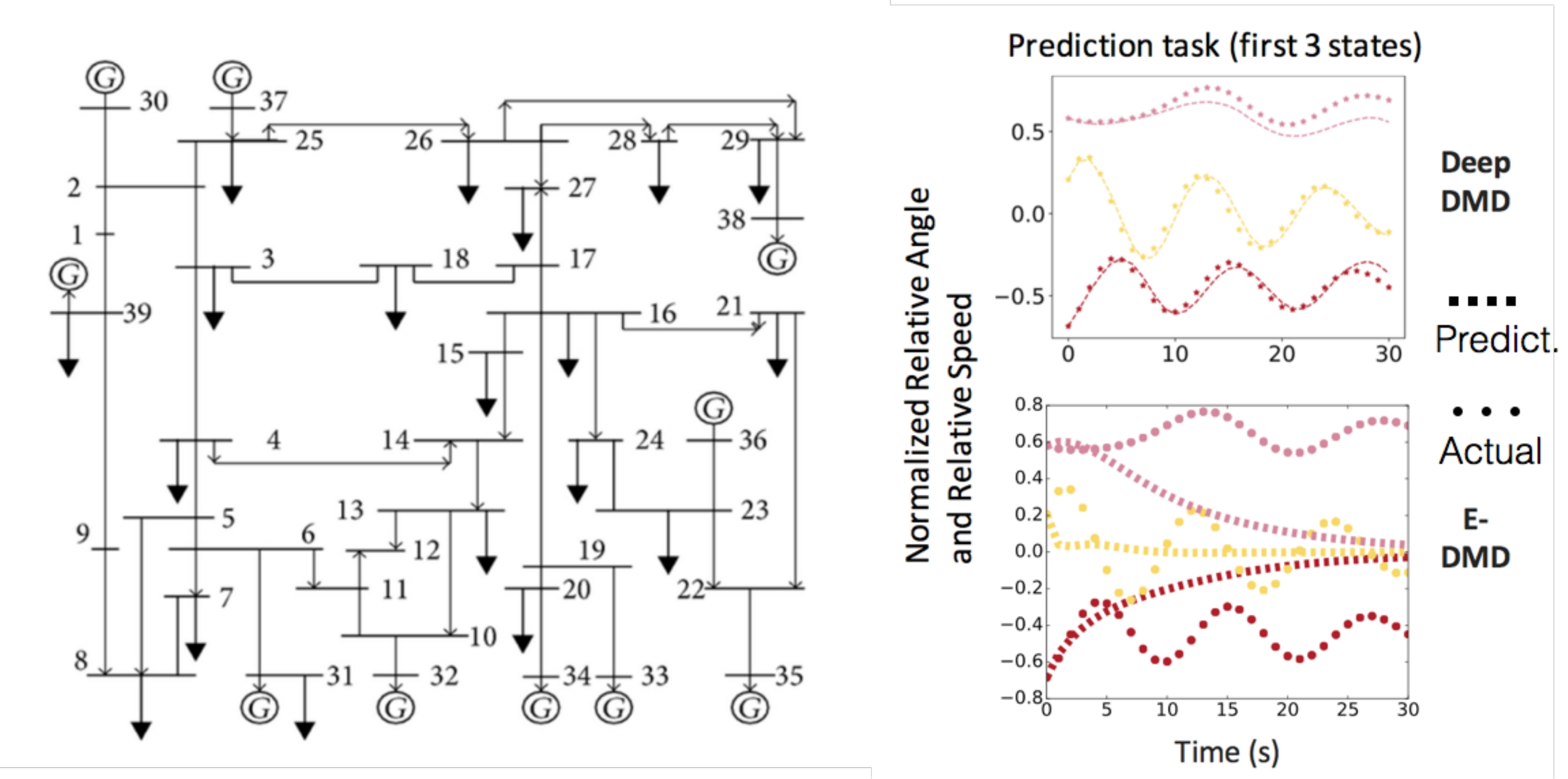}
\caption{(Left) The 10 generator IEEE 39-bus benchmark system. Circular nodes labeled G represent generators, buses are denoted by the intersection of two lines, individual lines represent transmission lines, and arrows represent loads on the network.  (Right) A plot of Koopman operator predictions for the relative angle and relative speed dynamics of the IEEE 39-bus benchmark system.  The top plot shows the prediction for the first three states of the system using deep dynamic mode decomposition and the bottom plot shows the prediction using extended dynamic mode decomposition.   }
\label{fig:Figure5}
\end{figure}

During deep dynamic mode decomposition, we selected the deep Koopman operator that minimized 1-step prediction training error.  At the test stage, we then generated a random initial condition different from any of the initial conditions used to generate training trajectories.  The outcomes are plotted in Figure \ref{fig:Figure5}.  Our benchmark again is multi-step prediction accuracy, given time-series data. 

We observed that the deep dynamic mode decomposition algorithm performed much better at multi-step prediction than Legendre-polynomial based dynamic mode decomposition. Again it is possible that with another choice of dictionary functions, e.g.  Hermite polynomials or thin-plate RBFs, we would obtain more accurate results.    The insight from these results is that the deep dynamic mode decomposition algorithm appears to be able to adaptively learn the Koopman basis required for both glycolytic and swing dynamic oscillations.  Notice that the oscillatory nature of the glycolysis dynamics are more of a relaxation-type oscillator, while the swing dynamics follow a damped sinusoidal response.  The deep dynamic decomposition algorithm appears to be able capture both types of nonlinearities, albeit with potentially different weighting parameters within the deep neural networks.  For the power system example, we found that a 20-layer ELU feedforward network was adequate to recapitulate the swing dynamics of the system.  However, Liao et al. showed that a shallow 3-layer neural network with Tikhonov regularization using the tanh activation function, can also learn oscillatory dynamics for  the Duffing oscillator \cite{Li2017}.      Future research will investigate ways to characterize the minimal depth, or neural net complexity, as well as the impact of different activation functions, required to learn the Koopman operator for different types of systems. 

\section{Conclusion}
In this paper we presented a deep learning approach to calculating the Koopman operator.  Our automated dictionary approach, enabled by deep learning, improves the performance of the Koopman operator for longer term forecasting. We compared it with the current state of the art methods, extended dynamic mode decomposition, and we showed that in many cases the deep Koopman operator performs relatively well at multistep prediction tasks.  Our results suggest that the deep Koopman operator provide a complementary and promising alternative to extended dynamic mode decomposition approaches for data-driven modeling problems of complex nonlinear systems.

\bibliographystyle{plain}

\begin{thebibliography}{2}
\bibitem{Koopman1931}Koopman, B. O. (1931). Hamiltonian systems and transformation in Hilbert space. Proceedings of the National Academy of Sciences, 17(5), 315-318.
\bibitem{Mezic2004}Mezic, I., \& Banaszuk, A. (2004). Comparison of systems with complex behavior. Physica D: Nonlinear Phenomena, 197(1), 101-133.
\bibitem{Mezic2005}Mezic, I. (2005). Spectral properties of dynamical systems, model reduction and decompositions. Nonlinear Dynamics, 41(1), 309-325.
\bibitem{Rowley2009}Rowley, C. W., Mezi?, I., Bagheri, S., Schlatter, P., \& Henningson, D. S. (2009). Spectral analysis of nonlinear flows. Journal of fluid mechanics, 641, 115-127.
 \bibitem{Williams2015}Williams, M. O.,\ \& Kevrekidis, I. G., \& Rowley, C. W. (2015). A data-driven approximation of the koopman operator: Extending dynamic mode decomposition. Journal of Nonlinear Science, 25(6), 1307-1346.
Chicago\\	
\bibitem{Brunton2016}Brunton, S. L., Proctor, J. L., \& Kutz, J. N. (2016). Discovering governing equations from data by sparse identification of nonlinear dynamical systems. Proceedings of the National Academy of Sciences, 113(15), 3932-3937.
Chicago	
\bibitem{Proctor2016}Proctor, J. L., Brunton, S. L., \& Kutz, J. N. (2016). Dynamic mode decomposition with control. SIAM Journal on Applied Dynamical Systems, 15(1), 142-161.\\ 
\bibitem{Li2017} Li, Q., Dietrich, F., Bollt, E. M., \& Kevrekidis, I. G. (2017). Extended dynamic mode decomposition with dictionary learning: a data-driven adaptive spectral decomposition of the Koopman operator. arXiv preprint arXiv:1707.00225.
\bibitem{Duchi2007}Duchi, J., Hazan, E., \& Singer, Y. (2011). Adaptive subgradient methods for online learning and stochastic optimization. Journal of Machine Learning Research, 12(Jul), 2121-2159.\\
\bibitem{Kingma2014}Kingma, D., \& Ba, J. (2014). Adam: A method for stochastic optimization. arXiv preprint arXiv:1412.6980.
\bibitem{Srivastava2014}Srivastava, N., Hinton, G. E., Krizhevsky, A., Sutskever, I., \& Salakhutdinov, R. (2014). Dropout: a simple way to prevent neural networks from overfitting. Journal of machine learning research, 15(1), 1929-1958.
\bibitem{Williams2016}Williams, M. O., Hemati M. S., Dawson S. T.,  Kevrekidis I. G. \& Rowley, C. W. (2016). Extending Data-Driven Koopman Analysis to Actuated Systems. IFAC-PapersOnLine, 49(18), 704-709. 
\bibitem{Raghu2016}Raghu, M., Poole, B., Kleinberg, J., Ganguli, S., \& Sohl-Dickstein, J. (2016). On the expressive power of deep neural networks. arXiv preprint arXiv:1606.05336.
\bibitem{Bengio2011}Bengio, Yoshua, and Olivier Delalleau. "On the expressive power of deep architectures." International Conference on Algorithmic Learning Theory. Springer Berlin Heidelberg, 2011.
\bibitem{Bengio2013}Bengio, Y., Mesnil, G., Dauphin, Y., \& Rifai, S. (2013). Better Mixing via Deep Representations. In ICML (1) (pp. 552-560).
\bibitem{Kowalski1991}Kowalski, K., \& Steeb, W. H. (1991). Nonlinear dynamical systems and Carleman linearization. World Scientific.
\bibitem{Goncalves2008} Gon\c alves, J., \& Warnick, S. (2008). Necessary and sufficient conditions for dynamical structure reconstruction of LTI networks. IEEE Transactions on Automatic Control, 53(7), 1670-1674.
\bibitem{Daniels2015}Daniels, B. C., \& Nemenman, I. (2015). Efficient inference of parsimonious phenomenological models of cellular dynamics using S-systems and alternating regression. PloS one, 10(3), e0119821.
\bibitem{Kundur1994}Kundur, P. (1994). Power system stability and control (Vol. 7). N. J. Balu, \& M. G. Lauby (Eds.). New York: McGraw-hill.
\bibitem{Grainger1994} Grainger, J. J. S., Grainger, W. D. J. J., \& Stevenson, W. D. (1994). Power system analysis. 
\bibitem{Fouad1991} Fouad, Abdel-Azia, and Vijay Vittal. Power system transient stability analysis using the transient energy function method. Pearson Education, 1991.
\bibitem{Huang2009} Huang, Z., Chen, Y., \& Nieplocha, J. (2009, July). Massive contingency analysis with high performance computing. In Power \& Energy Society General Meeting, 2009. PES'09. IEEE (pp. 1-8). IEEE.
\bibitem{MezicPower} Susuki, Y., \& Mezic, I. (2011). Nonlinear Koopman modes and coherency identification of coupled swing dynamics. IEEE Transactions on Power Systems, 26(4), 1894-1904.
\bibitem{MezicSpectral} Susuki, Y., \& Mezic, I. (2012). Nonlinear Koopman modes and a precursor to power system swing instabilities. IEEE Transactions on Power Systems, 27(3), 1182-1191.
\bibitem{MezicSpectralStability} Mauroy, A., \& Mezic, I. (2016). Global stability analysis using the eigenfunctions of the Koopman operator. IEEE Transactions on Automatic Control, 61(11), 3356-3369.
\end{thebibliography}

\end{document}